\pdfoutput=1

\documentclass[%
 aip,
 amsmath,amssymb,
 reprint,
]{revtex4-1}

\usepackage{graphicx}
\usepackage{dcolumn}
\usepackage{bm}
\usepackage{url}
\usepackage[utf8]{inputenc}
\usepackage[T1]{fontenc}
\usepackage{mathptmx}
\usepackage{etoolbox}

\makeatletter
\def\@email#1#2{%
 \endgroup
 \patchcmd{\titleblock@produce}
  {\frontmatter@RRAPformat}
  {\frontmatter@RRAPformat{\produce@RRAP{*#1\href{mailto:#2}{#2}}}\frontmatter@RRAPformat}
  {}{}
}%
\makeatother

\begin{document}

\preprint{AIP/123-QED}

\title[Hybrid-Domain Synergistic Transformer for Hyperspectral Image Denoising]{Hybrid-Domain Synergistic Transformer \\ for Hyperspectral Image Denoising}

\author{Haoyue Li}

\author{Di Wu}%
 \email{wudi@suda.edu.cn.}
\affiliation{ 
School of Optoelectronic Science and Engineering, Soochow University
}%

\date{\today}

\begin{abstract}
Hyperspectral image denoising faces the challenge of multi-dimensional coupling of spatially non-uniform noise and spectral correlation interference. Existing deep learning methods mostly focus on RGB images and struggle to effectively handle the unique spatial-spectral characteristics and complex noise distributions of hyperspectral images (HSI). This paper proposes an HSI denoising framework, Hybrid-Domain Synergistic Transformer Network (HDST), based on frequency domain enhancement and multiscale modeling, achieving three-dimensional collaborative processing of spatial, frequency and channel domains. The method innovatively integrates three key mechanisms: 1) introducing an FFT preprocessing module with multi-band convolution to extract cross-band correlations and decouple spectral noise components; 2) designing a dynamic cross-domain attention module that adaptively fuses spatial domain texture features and frequency domain noise priors through a learnable gating mechanism; 3) building a hierarchical architecture where shallow layers capture global noise statistics using multiscale atrous convolution, and deep layers achieve detail recovery through frequency domain postprocessing. Experiments on both real and synthetic datasets demonstrate that HDST significantly improves denoising performance while maintaining computational efficiency, validating the effectiveness of the proposed method. This research provides new insights and a universal framework for addressing complex noise coupling issues in HSI and other high-dimensional visual data. The code is available at \url{https://github.com/lhy-cn/HDST-HSIDenoise}.
\end{abstract}

\maketitle

\section{Introduction}

Image denoising, a traditional image processing technique, aims to recover clean information from observed signals and finds extensive applications in tasks such as remote sensing \cite{Yang2024,9949236}, target recognition \cite{9554871}, and 3D reconstruction \cite{Li2019ResearchOD}. Traditional methods such as wavelet transform \cite{6351286}, nonlocal means \cite{Kaur2015ComparisonON} rely on hand-crafted priors and are inadequate for handling complex noise distributions. In recent years, deep learning-based denoising models such as DnCNN \cite{zhao2018low}, SwinIR \cite{liang2021swinir} have significantly improved performance through powerful image feature extraction capabilities. These image denoising algorithms primarily focus on RGB images, which typically have only 3 channels and a simple structure. However, hyperspectral images (HSI), commonly used in fields such as agriculture and environmental monitoring \cite{8314827}, capture spectral-spatial information from scenes through hundreds of contiguous narrow-band channels. Traditional algorithms cannot be directly applied to HSI denoising due to their neglect of channel correlations \cite{8913713}. Compared to RGB images, HSI denoising presents the following unique challenges:

\textbf{First}, HSI exhibits coupled multidimensional noise and cross-band correlations. Noise is non-uniformly distributed across spatial dimensions and complexly coupled with spectral dimensions, exhibiting inter-band dependencies \cite{9855427}.

\textbf{Second}, sensitivity to high-frequency information: Inconsistent imaging equipment causes HSIs to frequently have bit depths different from RGB. Weak spectral features are easily drowned out by noise, requiring fine-grained frequency-domain guidance\cite{Yang2024}.

\textbf{Third}, detail preservation and computational efficiency: Hundreds of channels substantially increase computational complexity. Traditional methods lack efficiency and capability in jointly modeling local spatial details and global spectral continuity \cite{Ozdemir2020}.

In recent years, frequency domain analysis has shown unique advantages in image restoration \cite{Zhang_2024,zheng2025lightweightmedicalimagerestoration}. Successful applications in RGB image denoising include Fourier convolutions in FADformer \cite{10.1007/978-3-031-72940-9_14} and cross-domain filtering in FCENet \cite{Zhu_2021_CVPR}, but their adaptation to HSI remains to be explored. Additionally, Atrous Spatial Pyramid Pooling (ASPP) \cite{DBLP:journals/corr/ChenPK0Y16} captures noise components of different frequencies through multi-scale atrous convolutions, and its multi-scale feature extraction capability has been validated in tasks such as semantic segmentation \cite{LIAN2021107622}. This module is also widely used in image processing; for example, ASPP-DF-PVNet \cite{ZHU2021116268} improves the generalizability of the model by better capturing details at different levels. When frequency domain analysis is combined with ASPP, the multi-dilation rate design of ASPP can be mapped to multi-band processing in the frequency domain. This feature allows it to specifically separate the periodic spectrum of stripe noise, composed of low-frequency directional energy \cite{DBLP:journals/corr/ChenPK0Y16}, from the high-frequency random components of sensor noise in HSI denoising, providing frequency domain priors with clear physical meaning for subsequent cross-domain feature fusion. These advancements offer new ideas for HSI denoising, but integrating frequency domain analysis with the spatial-spectral dual characteristics of HSI requires further exploration.

To this end, this paper proposes an HSI denoising framework based on frequency domain enhancement and multiscale modeling, achieving spatial-frequency-channel three-dimensional collaborative processing. Specifically, building upon SERT \cite{li2023spectralenhancedrectangletransformer}, it innovatively deeply integrates frequency domain noise decoupling with spatial multi-scale modeling: 1) Introducing an fast Fourier transform (FFT) preprocessing module to transform HSI into the frequency domain for noise separation, combined with ASPP multi-scale atrous convolution \cite{DBLP:journals/corr/ChenPK0Y16} to suppress noise components of different frequencies; 2) Designing a dynamic cross-domain attention module to adaptively fuse spatial-frequency features through a learnable gating mechanism, such as enhancing noise suppression in low-frequency regions dominated by the frequency domain and preserving ground object textures in high-frequency regions dominated by the spatial domain; 3) Constructing a shallow global-deep local hierarchical architecture, utilizing the global average pooling branch of ASPP to capture noise statistical characteristics  and guide detail recovery in deep networks. Experiments show that this framework improves the quality of HSI restoration in mixed noise scenarios while maintaining the efficiency of SERT\cite{li2023spectralenhancedrectangletransformer} to a certain extent. This method provides a universal framework for denoising high-dimensional visual data such as HSI and multi-modal remote sensing images.

In general, our contributions can be summarized as follows:

1. Propose a spatial-frequency-channel three-dimensional collaborative dynamic cross-domain feature fusion framework, combining spatial domain Transformer with FFT-based frequency domain feature extraction in a post-processing manner. It realizes adaptive weight allocation of dual-domain features through a learnable gating mechanism, enabling cross-domain attention for dual-domain collaborative processing to address the coupling of HSI noise in spatial, spectral, and frequency dimensions.

2. Widely introduce spatial pyramid pooling into the joint processing of spatial and frequency domains. While constructing a hierarchical denoising chain of global statistical guidance-local detail recovery, it forms cross denoising capabilities of spatial multi-resolution + frequency multi-band.

3. Compared to the baseline, our model achieves a PSNR improvement of 0.94 dB on Realistic dataset and a maximum PSNR improvement of 0.52 dB on ICVL synthetic dataset.

\section{Related Work}

\subsection{Traditional HSI Denoising Based on Frequency Domain Processing}
Traditional HSI denoising methods based on frequency domain processing mainly utilize the Fourier transform and the wavelet transform \cite{Palsson2014HyperspectralID,rasti2018noise}. Fourier transform distinguishes noise and signal distributions through frequency domain conversion; for example, low-pass filters can suppress high-frequency noise but tend to lose image details and have limited effectiveness for complex noise. The wavelet transform decomposes HSI into sub-bands of different frequencies and achieves denoising by processing coefficients of different sub-bands \cite{6351286,zhou2009wavelet}. In certain sub-bands, if noise energy accounts for a large proportion, thresholding can be applied to the sub-band coefficients to suppress noise components. However, the selection of wavelet basis functions and the determination of optimal thresholds are critical issues for this method. Improper threshold selection can lead to over-denoising resulting in image blurring or under-denoising leaving significant noise \cite{zhou2009wavelet}. The aforementioned traditional methods mostly consider only the frequency domain perspective and do not fully integrate the spatial and spectral characteristics of HSI, making them unsuitable for current demands for high-quality HSI processing.

\subsection{Deep Learning Based HSI Denoising}
Early studies attempted to apply deep learning to HSI denoising. For example, Xie and Li \cite{8030333} first used deep CNNs with trainable nonlinear functions for HSI denoising. However, extending 2D models to 3D HSI faces challenges: under-utilization of spatial-spectral correlations, high computational complexity, and insufficient data \cite{Dong2019DeepSR}.
To address this, Dong et al. proposed a 3D HSI denoising framework using an improved 3D U-net\cite{Dong2019DeepSR}. It employs separable filtering that decomposes 3D filtering into 2D spatial and 1D spectral to reduce complexity and transfer learning to supplement data, outperforming model-based methods.
Recent innovations include: Liang et al.’s HSSD \cite{10.1007/978-3-031-91838-4_17}, which combines CNN and Transformer with a decoupling strategy to capture local / nonlocal details and global spectral correlations, achieving superior spatial-spectral reconstruction; Fu et al.’s GMSC-Net \cite{9553257}, a model-based method using sparse representation and iterative optimization, addressing the lack of clean-noisy pairs through clustering. These studies advance deep learning-based HSI denoising.

\begin{figure*}
\includegraphics[width=\textwidth]{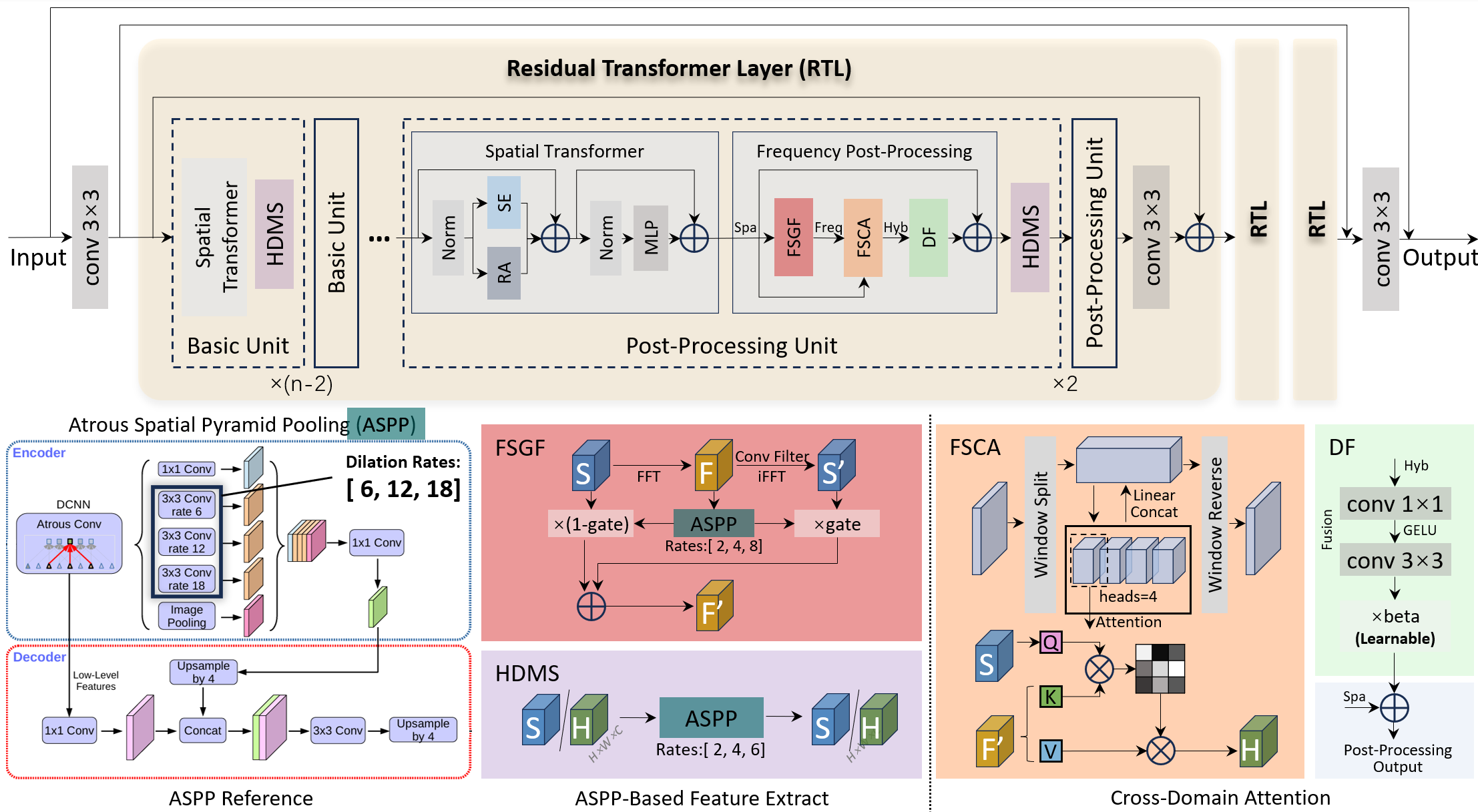}
\caption{\label{fig:net}Overall framework of HDST.}
\end{figure*}

\subsection{Deep Learning Image Denoising Combined with Frequency Domain}
Recent progress combines frequency domain with deep learning for image restoration \cite{}, leveraging frequency transformation to decouple noise and signal spectra, while utilizing the nonlinear modeling capabilities of deep networks to achieve efficient denoising. For example, Gao et al. proposed a frequency domain deraining method based on contrastive regularization to improve feature distinguishability\cite{gao2024efficient}. Kong et al. designed a Frequency-domain Self-Attention Solver (FSAS) and Discriminative Frequency-domain Feed-forward Network (DFFN) using the convolution theorem to improve deblurring efficiency\cite{kong2023efficient}. Wang et al. mined the frequency domain complementarity of NIR and RGB images through the FCENet framework to achieve adaptive fusion\cite{wang2025complementary}. These works have demonstrated the potential of frequency domain processing but require targeted optimization for HSI denoising. Meanwhile, cross-domain fusion frameworks have also been widely explored. For example, Zhang et al.'s FS-Net \cite{Zhang_2024} separates frequency components of underwater images and combines spatial domain filtering to enhance contrast. In the HSI domain, Wang et al.'s UOANet \cite{10149468} and Sheng et al.'s SANet \cite{Sheng_2023_CVPR} optimize denoising effects through spectrum-spatial feature fusion and frequency domain guided denoising.
Inspired by this, this paper proposes a closed-loop cross-domain fusion framework to improve the ability to distinguish between textures and noise through dynamic decoupling of frequency and spatial features.

\section{Method}

Spatial domain methods are widely used in image restoration and can generally achieve relatively good denoising levels\cite{10144690}. To enable our proposed frequency domain theoretical framework to be transplanted and applied to a large number of spatial domain denoising models, we integrate the idea of post-processing into the frequency domain theoretical framework while ensuring that the overall input and output of this framework maintain the same dimensional size, allowing information captured in the frequency domain to assist in improving the overall performance of the network.

The overall framework of the proposed method is shown in Fig.~\ref{fig:net}, which is based on the Spectral Enhanced Rectangle Transformer (SERT)\cite{li2023spectralenhancedrectangletransformer}. To further reduce high-frequency signals caused by noise, this paper proposes a frequency post-processing (FPP) module based on both frequency and spatial domains, along with a learnable gating mechanism to achieve adaptive weight allocation of dual-domain features.

\subsection{Review of SERT}

Spectral Enhanced Rectangle Transformer (SERT)\cite{li2023spectralenhancedrectangletransformer} employs 3 Residual Transformer Layers (RTL), each cascading 6 transformer blocks. Its core modules include: 1) Rectangular Attention (RA) module, which realizes multidirectional nonlocal interaction through horizontal and vertical rectangular window partitioning and spectral shuffling; 2) Spectral Enhancement (SE) module, which utilizes a global memory unit to store spectral priors and enhances spectral-spatial correlations while suppressing noise through a dynamic low-rank selection mechanism. These two modules work synergistically to effectively address the issues of multidimensional noise coupling and cross-band correlations in HSI, integrating the spatial and channel information of the image.

However, effective information in HSI is mostly concentrated in low- and medium-frequency regions, while random noise manifests itself as random fluctuations in pixel values in the spatial domain, and its energy is often widely distributed in high-frequency parts when transformed to the frequency domain. Therefore, SERT cannot effectively distinguish between stripe noise and edge textures in HSI space, leading to loss of image details after denoising.

\subsection{Hybrid-Domain Synergistic Transformer Network (HDST)}

\begin{figure}[t!]
\includegraphics[width=5cm,angle=90]{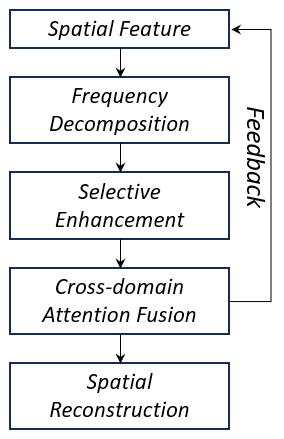} 
\caption{\label{fig:crossdomain} Cross-domain feedback loop}
\end{figure}

This framework takes RTL as the basic skeleton and innovatively introduces a frequency domain post-processing unit and a hybrid domain multiscale module, constructing a closed-loop optimization process of "spatial feature extraction - frequency domain noise decoupling - cross-domain information calibration". The loop is shown in Fig.~\ref{fig:crossdomain}

Specifically, hierarchical domain adaptation is performed for the cascaded Transformer structure within RTL. Shallow Transformers capture long-range spatial dependencies and low-rank spectral characteristics through the RA and SE modules, laying the foundation for feature representation; FPP units are embedded in the deepest two layers of Transformers to decouple noise spectral components through frequency domain transformation, avoiding blurring of shallow details due to premature frequency domain processing.
Within the FPP unit, through a three-stage chain of frequency domain multiscale filtering - cross-domain feature calibration - dynamic residual fusion, frequency domain noise patterns are converted into guiding signals for spatial domain denoising. Through the frequency-spatial collaborative attention mechanism, frequency domain noise priors such as specific frequency noise patterns are fed back to spatial domain feature calibration, and then adaptive control of feature enhancement is achieved through learnable residual connections, ultimately forming a complete optimization chain of "noise decoupling - feature calibration - error compensation".
Meanwhile, multiscale cross-enhancement is widely used to improve feature extraction capability. After each Transformer block, the HDMS module is used for multi-scale sampling, covering multilevel information from edge details to global structures, providing a richer feature base for frequency domain processing; ASPP-FFT located within FSGF realizes refined separation of wide-band noise in the frequency domain through multi-dilation rate atrous convolution, effectively achieving two-dimensional multiscale analysis.

\subsubsection{Atrous Spatial Pyramid Pooling (ASPP)}

\textbf{ASPP} Atrous Spatial Pyramid Pooling is a module proposed in the DeepLab system \cite{DBLP:journals/corr/ChenPK0Y16}. It contains 5 parallel branches: 

\begin{equation*}
\begin{cases}
1\times1\ \text{convolution} & \text{(preserve resolution)} \\
\text{Dilated conv (rates } d=6,12,18) & \text{(multiscale context)} \\
\text{Global Avg Pool} \xrightarrow[]{\text{1x1 Conv}} \text{Upsample} & \text{(global statistics)}
\end{cases}
\end{equation*}
The features are concatenated and fused by \(1\times1\) convolution. The structure is also shown in the ASPP Reference section of Fig.~\ref{fig:net}, with dilation rates [6,12,18]. 

\textbf{ASPP-FFT} Innovatively, we adapt this spatial domain module for frequency domain processing by employing multiscale frequency domain convolutions with dilation rates [2,4,8]. This enables the extraction of both local frequency-band correlations and cross-band global contextual information, which subsequently drives the generation of frequency-guided spatial gating signals. Since the frequency domain inherently represents global information, the redundant global statistics branch is removed to avoid disturbance.

\subsubsection{FFT-Scale Gated Fusion (FSGF)}

We perform FFT on the input spatial feature $S$, separate the real and imaginary parts, and concatenate them into $F_c$. Multiscale feature extraction in the frequency domain is achieved through the ASPP-FFT module. This enables wide-band noise filtering in the frequency domain, forming spatial-frequency dual-dimensional multiscale analysis. Its forward propagation can be expressed as:

\begin{equation}
F = FFT(S), F_c = Concat(F.real, F.imag)
\end{equation}
\begin{equation}
F_{proc} = ASPP\mbox{-}FFT(F_c, dilation = [2,4,8])
\end{equation}

where $S$ represents the input spatial domain information, $F$ is the frequency domain result obtained after a fast Fourier transform, and the feature $F_c$, formed by concatenating the real and imaginary parts of $F$, is used for multiscale ASPP-based analysis.

Subsequently, a dynamic gating mechanism generates a spectral mask $Gate$, which adaptively selects the fusion path based on noise intensity. The hyperparameter $\alpha$ is used to control the injection intensity of the information from the frequency domain. This process is expressed by the formulas:

\begin{equation}
S' = Conv_{3 \times 3}(IFFT(F_{proc}))
\end{equation}
\begin{equation}
Gate = \sigma(Conv_{1 \times 1}(F_{proc}))
\end{equation}
\begin{equation}
F' = S \odot Gate + \alpha \cdot S' \odot (1 - Gate)
\end{equation}
where $S'$ is the spatial feature reconstructed from the frequency domain processing result, \(3\times3\) convolution denotes the reconstruction convolution operation;  $\odot$ represents element-wise multiplication; Gate retains clean areas (\(\text{Gate} \to 1\)) and replaces noisy regions (\(\text{Gate} \to 0\)).

\subsubsection{Frequency-Spatial Collaborative Attention (FSCA)}

The frequency domain fusion features output by FSGF and the original input features are divided into non-overlapping windows, and cross-domain associations are established through a multi-head attention mechanism: using spatial features as queries ($Q$) and frequency domain features as keys ($K$) and values ($V$), realizing the interaction logic of spatial positions querying their corresponding frequency domain noise patterns. The spatial map $S$ and the map $F'$ generated by combining original and reconstructed features with the mask are divided into M×M non-overlapping windows and flattened to reduce computational complexity. Then, unbiased multi-head attention calculation is performed on the flattened sequences, expressed as:
\begin{equation}
Q = S_{flat} W_Q^T, K = F_{flat} W_K^T, V = F_{flat} W_V^T
\end{equation}
\begin{equation}
X_i = X[:, :, iC_h : (i+1)C_h], X \in \{Q, K, V\}
\end{equation}
\begin{equation}
AttnHead_i = Softmax\left(\frac{Q_i \cdot K_i^T}{\sqrt{C_h}}\right) \cdot V_i
\end{equation}
\begin{equation}
H = Concat(AttnHead_i) \cdot W_O^T
\end{equation}
where $W_Q$, $W_K$, $W_V$, and $W_O$ are learnable weight matrices, projecting input features to query, key, value, and output spaces; $C_h$ represents the dimension of a single attention head, and the multi-head attention mechanism is implemented by dividing features by dimension; $Q_i$, $K_i$, $V_i$ represent the query, key, and value vectors of the i-th head.

\subsubsection{Dynamic Fusion}

Residual calculation is performed on the calibrated feature $H$ output by FSCA through a learnable parameter $\beta$ and a lightweight convolution fusion network, achieving dynamic fusion of "original features + frequency domain enhanced residuals". Its forward propagation can be expressed as:
\begin{equation}
H_{fused} = Conv_{3 \times 3}(GELU(Conv_{1 \times 1}(H)))
\end{equation}
\begin{equation}
Output = S + \beta \cdot H_{fused}
\end{equation}
where $H_fused$ is the residual output enhanced by the fusion network. \(3\times3\) convolution restores dimensions and introduces spatial context information and continuity constraints, avoiding block effects that may arise from frequency domain reconstruction; $\beta$ is a learnable fusion coefficient initialized to 0.1, ensuring stability in the early training stage and intensity of enhancement after convergence, preventing overshoot of features or loss of information.

\subsubsection{Hybrid-Domain Multi-Scale Module (HDMS)}

As a basic component throughout each RTL layer, the HDMS module performs multiscale atrous convolution with dilation rates [2,4,8] on spatial features after each Transformer block. Small dilation rates capture edge and texture details, and larger dilation rates aggregate target structures and globally associated features. Through multi-branch parallel sampling, it compensates for the deficiency of single-scale convolution in handling complex noise, providing more stable feature input for subsequent frequency domain processing and significantly enhancing feature robustness. 

Shallow HDMS strengthens the noise resistance of local details, while deep HDMS provides spatial feature bases containing multiscale semantics for FPP units, enabling frequency domain noise decoupling to more accurately adapt to content structures at different levels, achieving cross-layer information paving.

\section{Experiments}
To evaluate the proposed HDST framework, experiments are conducted against model-based and deep-learning hyperspectral denoising methods. Further ablation studies and computational efficiency analyzes are performed to validate the contribution of key components and the practical viability of HDST.

Today, deep learning denoising combined with the frequency domain has been involved in RGB image processing\cite{Archana2024}. However, HSI denoising methods that integrate deep learning with frequency domain processing are relatively rare. When applying related RGB image processing methods to HSI image denoising tasks, the increase in the number of channels will lead to potential performance degradation. Therefore, the method of comparative experiments is still based on comparison with commonly used hyperspectral denoising methods.

The comparison methods include several traditional model-based hyperspectral image denoising methods, such as filter-based method (Block-Matching 4D algorithm (BM4D \cite{maggioni2012nonlocal})), orthogonal basis-based method (Non-local Graph Matching (NGMeet \cite{he2019non})), wavelet-based method(3D wavelets\cite{8077589}); also included are three deep learning-based methods, namely Hyperspectral Image Denoising Convolutional Neural Network (HSID-CNN \cite{8454887}), Multi-Attention Fusion Network (MAC-Net \cite{xiong2021mac}), and a relatively new spatial-spectral recurrent transformer U-Net (SSRT-UNet \cite{10463066}). 

\textbf{Implementation.} The proposed model is implemented using Pytorch 2.4.1, and experiments are carried out on a server equipped with an Intel Xeon Processor CPU and a Nvidia A100-PCIE-40GB running Ubuntu 22.04.

\textbf{Metrics.} \textbf{PSNR} \cite{sara2019image} focuses on pixel-level errors, calculating the mean square error between the original and distorted images, comparing it with the square of the maximum signal value, and expressing quality in decibels (larger values mean less distortion). \textbf{SSIM} \cite{sara2019image} emphasizes image structure and perceptual quality, evaluating similarity via structure, luminance, and contrast. Its evaluation results are more consistent with human visual perception. Ranging 0–1, values closer to 1 indicate higher quality. \textbf{SAM} \cite{Ozdemir2020} focuses on spectral similarity, measuring it via the angle between spectral vectors; smaller angles mean more similar spectra and better quality.

\textbf{Loss Function.} This paper introduces the L2 loss function \cite{terven2025comprehensive} for network training. Let the noisy data be x and the corresponding ground truth be y; the loss function can be expressed as:
\begin{equation}
\arg\min_{f_\theta} \|f_\theta(x) - y\|_2^2
\end{equation}

\subsection{Performance on Realistic Dataset}

For Realistic dataset\cite{8894531}, our experimental method is generally consistent with the training of the original SERT\cite{li2023spectralenhancedrectangletransformer} network. The data set contains 59 noisy hyperspectral images and their corresponding clean noise-free hyperspectral images. We followed the SERT division method, using the same 44 images and 15 images for training and testing. Each hyperspectral image channel contains 34 bands, ranging from 400 to 700 nm, and the image size within a single channel is 696×520. Overlapping 128×128 spatial regions are cropped and data augmentation is performed for training\cite{8894531}. Sufficient training can be achieved by setting the number of training epochs to 500. The learning rate is set to $1e-4$ for the first 200 epochs, changed to $5e-5$ at 200 epochs, and set to $1e-5$ at 400 epochs.

\begin{table}
\caption{\label{tab:realdata} Average results of different denoising methods on Realistic dataset}
\begin{ruledtabular}
\begin{tabular}{lcccc}
Method Category & Method & PSNR & SSIM & SAM \\
\hline
 & Noisy & 23.26 & 0.7609 & 17.329 \\
Traditional & BM4D\cite{maggioni2012nonlocal} & 29.04 & 0.9471 & 3.087 \\
 & Wavelet\cite{8077589} & 28.05 & 0.9046 & 5.85 \\
 & NGMeet\cite{he2019non} & 28.72 & 0.9511 & 2.735 \\
Deep Learning & HSID-CNN\cite{8454887} & 26.44 & 0.8992 & 5.242 \\
 & MAC-Net\cite{xiong2021mac} & 29.2 & 0.9489 & 4.099 \\
 & SSRT-UNet\cite{10463066} & 30.17 & 0.9549 & 2.506 \\
 & SERT\cite{li2023spectralenhancedrectangletransformer} (Baseline) & 29.68 & 0.9533 & 2.536 \\
 & \textbf{HDST (ours)} & \textbf{30.62} & \textbf{0.9555} & \textbf{2.417} \\
\end{tabular}
\end{ruledtabular}
\end{table}

\begin{figure*}
\includegraphics[width=\textwidth]{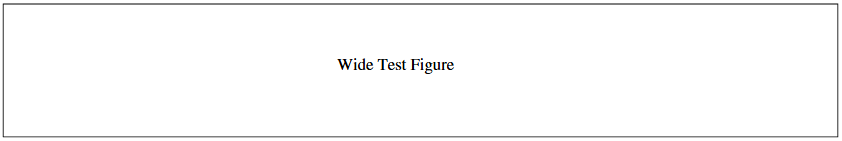}
\caption{\label{fig:data1}Visualization of denoising results on Realistic dataset. \\ \textbf{We are working on this figure, since we want to provide a more level of detail comparison for the visual comparison.}}
\end{figure*}

Table~\ref{tab:realdata} shows the average results of different methods on Realistic dataset. Our proposed HDST method significantly outperforms SERT\cite{li2023spectralenhancedrectangletransformer} as the baseline and several other hyperspectral image denoising methods, with a maximum PSNR improvement of 0.94dB, indicating the effectiveness of our method in handling real noise. In addition, we show the denoising results of the above real noisy hyperspectral images in Fig.~\ref{fig:data1}. In terms of noise removal and detail preservation, the results of our improved method are superior to traditional denoising methods and deep learning methods.

\subsection{Performance on ICVL Synthetic Dataset}

For ICVL synthetic dataset\cite{7afac8f318f34d9b81bc054068d8defc,bodrito2021trainablespectralspatialsparsecoding}, which contains 201 HSIs with a size of 1392×1300, and each hyperspectral image channel contains 31 bands, ranging from 400 to 700 nm. We followed the SERT\cite{li2023spectralenhancedrectangletransformer} division method, using the same 100 images, 5 images, and 50 images for training, validation, and testing. Training images are cropped to a size of 64×64 at different ratios. In the testing phase, HSI is cropped to 512×512×31 to obtain affordable computational costs. The number of training epochs is set at 100, with the learning rate set to $1e-4$ for the first 50 epochs and changed to $1e-5$ for the last 50 epochs.

The noise setting method partially follows the baseline settings, using the noise patterns in the literature \cite{7afac8f318f34d9b81bc054068d8defc} for simulation. The noise patterns are non-independently and identically distributed Gaussian noise (Non-i.i.d Gaussian noise), Gaussian + Stripe noise, Gaussian + Deadline noise, Gaussian + Impulse noise, and Mixture noise which contain stripe, deadline, impulse noise.

\begin{figure*}
\includegraphics[width=\textwidth]{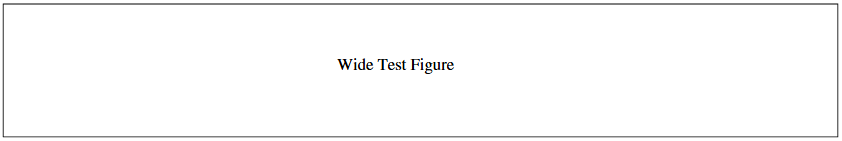}
\caption{\label{fig:data2}Visualization of denoising results on ICVL synthetic dataset.\\ \textbf{We are working on this figure, since we want to provide a more level of detail comparison for the visual comparison.}}
\end{figure*}

Table~\ref{tab:table3} presents the comparison results. The proposed HDST method also outperforms SERT\cite{li2023spectralenhancedrectangletransformer} as the baseline among the selected noise methods, and in most cases, it is completely superior to other denoising methods and performs more prominently on mixed noise approaching real situations; Fig.~\ref{fig:data2} presents the visualized comparison results. Our improved method has achieved relatively ideal visual effects, significantly outperforming denoising methods except SSRT-Unet, and slightly outperforming the newly proposed denoising model in recent years with fewer noise points in some areas.

\begin{table*}
\caption{\label{tab:table3}Average results of different denoising methods on various noise types in the synthetic dataset}
\begin{ruledtabular}
\begin{tabular}{cccccccccc}
Noise Pattern & \multicolumn{3}{c}{Gaussian+Mixture} & \multicolumn{3}{c}{Gaussian+Deadline} & \multicolumn{3}{c}{Gaussian+Impulse} \\
Method & PSNR & SSIM & SAM & PSNR & SSIM & SAM & PSNR & SSIM & SAM \\
\hline
Noisy & 13.91 & 0.3396 & 51.53 & 17.5 & 0.477 & 47.55 & 14.93 & 0.3758 & 46.98 \\
BM4D\cite{maggioni2012nonlocal} & 28.01 & 0.8419 & 23.59 & 33.77 & 0.9615 & 6.85 & 29.79 & 0.8613 & 21.59 \\
Wavelet\cite{8077589} & 26.09 & 0.7945 & 32.83 & 32.57 & 0.9476 & 8.877 & 28.87 & 0.8856 & 17.59 \\
NGMeet\cite{he2019non} & 26.13 & 0.7796 & 31.89 & 33.41 & 0.9665 & 6.55 & 27.02 & 0.7884 & 31.2 \\
HSID-CNN\cite{8454887} & 35.3 & 0.9588 & 6.29 & 38.33 & 0.9783 & 3.99 & 36.21 & 0.9663 & 5.48 \\
MAC-Net\cite{xiong2021mac} & 30.59 & 0.93 & 14.51 & 36.68 & 0.986 & 5.63 & 34.54 & 0.9553 & 10.2 \\
SSRT-UNet\cite{10463066} & 40.17 & 0.9902 & 2.968 & \textbf{44.01} & \textbf{0.996} & \textbf{1.661} & 42.59 & 0.9932 & 2.178 \\
SERT\cite{li2023spectralenhancedrectangletransformer} (Baseline) & 40.00 & 0.9937 & 2.84 & 43.66 & 0.9969 & 1.99 & 42.67 & 0.9959 & 2.3 \\
\textbf{HDST (ours)} & \textbf{40.52} & \textbf{0.9947} & \textbf{2.51} & 43.81 & 0.9966 & 1.83 & \textbf{42.89} & \textbf{0.9963} & \textbf{1.96} \\
\hline
\hline
Noise Pattern & \multicolumn{3}{c}{Gaussian+Stripe} & \multicolumn{3}{c}{Non-i.i.d Gaussian} & \multicolumn{3}{c}{ } \\
Method & PSNR & SSIM & SAM & PSNR & SSIM & SAM &   &   &   \\
\hline
Noisy & 17.51 & 0.4867 & 46.98 & 18.29 & 0.5116 & 46.2 &   &   &   \\
BM4D\cite{maggioni2012nonlocal} & 35.63 & 0.973 & 6.26 & 36.18 & 0.9767 & 5.78 &   &   &   \\
Wavelet\cite{8077589} & 34.17 & 0.9648 & 6.737 & 34.29 & 0.9658 & 6.623 &   &   &   \\
NGMeet\cite{he2019non} & 34.88 & 0.9665 & 5.42 & 34.9 & 0.9745 & 5.37 &   &   &   \\
HSID-CNN\cite{8454887} & 38.09 & 0.9765 & 4.59 & 39.28 & 0.9819 & 3.8 &   &   &   \\
MAC-Net\cite{xiong2021mac} & 39.03 & 0.991 & 4.03 & 39.98 & 0.9662 & 4.55 &   &   &   \\
SSRT-UNet\cite{10463066} & 43.75 & 0.9953 & 1.794 & 44.05 & 0.9962 & 1.691 & \textbf{ } & \textbf{ } & \textbf{ } \\
SERT\cite{li2023spectralenhancedrectangletransformer} (Baseline) & 43.68 & 0.9969 & 1.97 & 44.20 & 0.9971 & 1.69 &   &   &   \\
\textbf{HDST (ours)} & \textbf{43.88} & \textbf{0.9963} & \textbf{1.76} & \textbf{44.37} & \textbf{0.997} & \textbf{1.62} &   &   &   \\
\end{tabular}
\end{ruledtabular}
\end{table*}

\subsection{Ablation Study and Analysis}

To verify the theoretical analysis and understand the actual role of modules, in addition to the baseline and the proposed model, we additionally set up four groups of comparative models for experiments. Comparative Network 1 only adds frequency domain processing (FSGF+FSCA) to test the independent role of the frequency domain module. Comparative Network 2 adds Dynamic Fusion based on Network 1 to test the tuning ability of residual gating. Comparative Network 3 only adds multiscale pyramids (HDMS) to test the collaboration between multiscale and frequency domain ASPP. Comparative Network 4 includes frequency domain processing(without Dynamic Fusion) + multiscale pyramids(ASPP-FFT and HDMS)  to analyze the bottleneck of cross-domain feature fusion and assist in verifying the collaborative role of modules. The performance results of each comparative network in terms of PSNR and SSIM on Realistic dataset and its corresponding training environment are shown in Table~\ref{tab:ablation}.

\begin{table}
\caption{Ablation study on Realistic dataset}
\label{tab:ablation}
\begin{ruledtabular}
\begin{tabular}{lcccc}
Model & Frequency & Residual Fusion & ASPP/HDMS & PSNR \\
\hline
Baseline\cite{li2023spectralenhancedrectangletransformer} &  &  &  & 29.68 \\
Net1 & $\checkmark$ &  &  & 29.73 \\
Net2 & $\checkmark$ & $\checkmark$ &  & 29.89 \\
Net3 &  &  & $\checkmark$ & 30.27 \\
Net4 & $\checkmark$ &  & $\checkmark$ & 30.46 \\
HDST & $\checkmark$ & $\checkmark$ & $\checkmark$ & 30.62 \\
\end{tabular}
\end{ruledtabular}
\end{table}

The experimental results of Net1 and Net2 show that the processing of the independent frequency domain module may destroy spatial continuity, leading to a slight increase in parameters such as PSNR; while under the correction of the dynamic fusion module, the frequency domain module can achieve effective improvement; the experimental results of Net3 show that the pure multi-scale module brings the large single-module gain, verifying that multi-scale modeling is a key pillar for performance improvement; the experimental results of Net4 show that the combination of frequency domain and multi-scale methods achieves improvement, proving that multi-scale feature extraction provides a frequency spectrum decomposition basis with clear physical meaning for frequency domain processing. 

Both theoretical analysis and the above experiments show that multiscale methods can effectively collaborate with frequency domain methods, and the design of each module in the improved model is reasonable.

\subsection{Model Computational Efficiency Analysis}

We conducted statistics on the number of parameters and the operation speed on Realistic dataset and its corresponding training environment, and the statistical results are shown in Table~\ref{tab:complexity}. During the upgrade of the baseline structure, we observed that the number of parameters increased from 1.91M to 3.08M, an increase of 61\%, while the amount of computation (GFLOPS) increased only approximately 14\%. 

The core reason is that the newly added modules adopt a parameter-intensive but computationally efficient design strategy, achieving decoupling of the two. The multi-scale method uses multi-branch 1×1 convolution to build a spatial pyramid, whose parameters are concentrated in the channel projection layer, but the amount of computation is compressed through depth-wise separable convolution; the frequency domain processing module introduces fully connected layer parameters through real and imaginary part separation and dynamic gating, but controls the computational increment using the zero-parameter characteristics of the FFT transform and dimensionality reduction strategy. Finally, the cross-domain attention mechanism limits the attention range through window partitioning.

The asymmetric growth of parameters and computation reflects the idea of "trading parameters for efficiency", that is, enhancing the model's expressive ability by adding lightweight parameter modules and suppressing computational overhead through optimization, improving model capacity while maintaining computational efficiency.

\begin{table}
\caption{Model complexity analysis on Realistic dataset}
\label{tab:complexity}
\begin{ruledtabular}
\begin{tabular}{lrrr}
Metric & Baseline\cite{li2023spectralenhancedrectangletransformer} & HDST & $\Delta$(\%) \\
\hline
GFLOPS & 4086.2 & 4686.8 & +14.7 \\
Params (M) & 1.91 & 3.08 & +61.2 \\
\end{tabular}
\end{ruledtabular}
\end{table}

\subsection{Analysis and Discussion}

When comparing the improvement effect of the proposed model over the baseline on different datasets, it can be found that the improvement effect of HDST on Realistic dataset is overall better than that on ICVL synthetic dataset; among various noise methods in the synthetic dataset, the improvement effect of HDST on mixed noise is relatively the best. This phenomenon needs to be analyzed in depth from the perspective of the synergistic effect between data noise characteristics and network frequency domain processing mechanisms.

Real HSI noise has the characteristics of spatial non-uniformity and frequency domain energy aggregation. The frequency domain processing module of HDST fully utilizes this characteristic: FSGF constructs cross-band filters through multi-dilation rate convolution, which can directionally capture frequency noise patterns; and the gating mechanism triggers the frequency domain reconstruction path in noise regions to suppress noise. In mixed noise scenarios, the advantages of HDST are more significant, which benefits from the synergistic effect of multiscale frequency-spatial features. The shallow HDMS module extracts edge features through multi-scale atrous convolution, providing rich structured priors for frequency domain processing; the FSCA module uses spatial features as query signals to locate and suppress noise components from frequency domain features. This cross-domain collaboration mechanism enables HDST to maintain high denoising performance even in complex noise environments.

In contrast, the performance improvement of the synthetic dataset is relatively limited, possibly because its noise distribution is different from the real scenarios. Independent and identically distributed Gaussian noise and other synthetic noises are more uniformly distributed in the frequency domain, reducing the selectivity of the FSGF gating mechanism. In addition, although the sample size of the synthetic dataset is large, the noise patterns are relatively single, while Realistic dataset, although the number of samples is similar, has more diverse noise patterns, enabling the network to learn more generalized features.

In summary, the excellent performance of HDST on real data stems from the matching between the frequency domain separability of noise and the network's dual-domain collaboration mechanism, which also effectively demonstrates the rationality of our model design. The spatial non-uniformity and frequency domain energy aggregation characteristics of real noise enable the improved model to maximize its efficiency, while the mixed noise scenario verifies the cascaded enhancement effect of multiscale features and frequency domain processing. The uniform noise distribution and limited training scale of synthetic data constitute performance bottlenecks, indicating that future research should pay more attention to the authenticity modeling of noise physical models rather than relying solely on the generalization ability of synthetic data. This analysis provides important design criteria for frequency-domain-driven HSI denoising: priority should be given to the physical frequency-domain attributes of noise to achieve better denoising effects.

\section{Conclusion}

Hyperspectral image denoising needs to simultaneously address spatially non-uniform noise and spectral correlation interference. Traditional methods ignore the inherent multidimensional noise coupling and cross-band correlations of HSI images, leading to the limitation of single-domain processing being unable to balance detail preservation and noise suppression. This framework takes residual Transformer layers as the basic skeleton and constructs a closed-loop optimization process of "spatial feature extraction - frequency domain noise decoupling - cross-domain information calibration" through cross-layer collaboration between frequency domain post-processing units and hybrid domain multi-scale modules. This work demonstrates that hybrid domain learning is not merely algorithmic fusion but creates emergent properties: the closed-loop spatial-frequency interaction enables noise component transferability that single-domain methods cannot achieve. A large number of experimental results also prove the effectiveness of the proposed method, which has certain possibilities for practical application.

\begin{acknowledgments}
We acknowledge that there is no funding source that supports this research. 

This preprint uses a template adapted from AIP Advances to improve readability. This is not the target journal I intend to submit to; the template was initially adopted simply because I once tried to submit other articles using it. The final version will conform to the format requirements of the target journal.

This preprint reports work-in-progress. Full ablation studies and refined technical descriptions will be updated in subsequent versions. Current conclusions focus on core methodological validity.
\end{acknowledgments}

\section*{Data Availability Statement}

The data that support the findings of this study are available from the Realistic dataset: \url{https://github.com/ColinTaoZhang/HSIDwRD} and the ICVL dataset: \url{https://huggingface.co/datasets/danaroth/icvl}. The code implementing the proposed HDST is openly available on GitHub at \url{https://github.com/lhy-cn/HDST-HSIDenoise}.

\bibliography{HDST}

\end{document}